\documentclass[a4paper,conference]{IEEEtran}

\usepackage{graphicx} 
\usepackage{subfigure}
\usepackage{multirow}
\usepackage{makecell}

\ifCLASSINFOpdf
\else
\fi
\begin{document}
%
\title{CG-DIQA: No-reference Document Image Quality Assessment Based on Character Gradient}

\author{Hongyu~Li
	and~Fan~Zhu
	\thanks{H. Li and F. Zhu are with the AI Lab
		of ZhongAn Information Technology Service Co., Ltd, Shanghai,
		200002 China e-mail: lihongyu@zhongan.io.}
	\thanks{Manuscript received XXX XX, 201X; revised XXX XX, 201X.}}

\author{\IEEEauthorblockN{Hongyu Li}
\IEEEauthorblockA{AI Lab\\
ZhongAn Information\\ Technology Service Co., Ltd.\\
Shanghai, China\\
Email: lihongyu@zhongan.io}
\and
\IEEEauthorblockN{Fan Zhu}
\IEEEauthorblockA{AI Lab\\
	ZhongAn Information\\ Technology Service Co., Ltd.\\
	Shanghai, China\\
	Email: zhufan@zhongan.io}
\and
\IEEEauthorblockN{Junhua Qiu}
\IEEEauthorblockA{AI Lab\\
	ZhongAn Information\\ Technology Service Co., Ltd.\\
	Shanghai, China\\
	Email: qiujunhua@zhongan.io}}


%


\maketitle

\begin{abstract}
Document image quality assessment (DIQA)  is an important and challenging problem in real applications.
In order to predict the quality scores of document images, this paper proposes a novel no-reference DIQA
method based on character gradient, where the OCR accuracy is used as a ground-truth quality metric. Character gradient is computed on character patches detected with the maximally stable extremal regions (MSER) based method. Character patches are essentially significant to character recognition and therefore suitable for use in estimating document image quality. Experiments on a benchmark dataset show that the proposed method outperforms the state-of-the-art methods in estimating the quality score of document images.
\end{abstract}


\IEEEpeerreviewmaketitle

\section{Introduction}

\IEEEPARstart{W}{ith} the pervasive use of smartphones in our daily life, acquring document images with mobiles is becoming popular in digitization of business processes.
The optical character recognition (OCR) performance of mobile captured document images is often decreased with the low quality due to artifacts introduced during image
acquisition \cite{Ye2013Document}, which probably hinders the following business process severely.  For example, during online insurance claims, if a document image of low quality, submitted for claims, is not detected as soon as possible to require a recapture, 
critical information may be lost in business processes once the document is unavailabe later. 
To avoid such information loss, therefore, automatic
document image quality assessment (DIQA) is necessary and of great value in document processing and analysis tasks.

Methods for natural image quality assessment may not be suitable for
document images because both the properties
of document images and the objective of DIQA are totally different. To estimate the quality of document images, many no-reference (NR)
assessment algorithms have been proposed, where the reference document
image is not accessible in most practical cases.

According to the difference of feature extraction, these NR DIQA methods can be categorized as two groups: learning-based assessment and metric-based 
assessment.

The learning-based DIQA methods take advantage of learning techniques, such as  deep learning \cite{Kang2014A}, to extract discriminant features for different types of document degradations. 
They perform well only on the dataset on which they were trained. However, it is unrealistic to collect sufficient document samples for training in real applications.

The metric-based methods usually are based on hand-crafted features that are correlated with the OCR accuracy.Some degradation-specific quality metrics have been proposed to measure noise and character shape preservation \cite{Nayef2015Metric}.   
Although much progress has been made in metric-based assessment, there still exists a clear problem. Features used in existing methods are generally extracted from square image patches, many of which do not have visual meaning involving character/text. Therefore,  the
resultant features, probably containing much noise, are not optimal for DIQA.

\begin{figure}
	\centering   
	\includegraphics [width=0.50\textwidth]{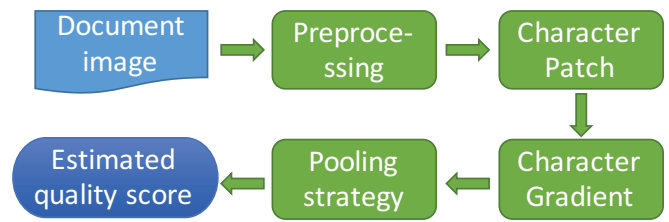}  
	\caption{Flowchart of the CG-DIQA method} 	
	\label{fig:overview}
\end{figure}

In addition, although the document
image quality is affected by many degradations caused during image acquisition, blur is often considered as the most common issue in mobile captured document images, arising from defocus, camera motion, or hand shake \cite{Ye2013Document,Rusinol2014Combining,Chabard2015Local}.  Moreover, the blur degradation has a bad impact on the OCR accuracy, which suggests that detecting the blur degradation is more attractive and useful in practical applications. 

The most striking feature distinguishing document
images from other types of images is character/text. As a consequence, DIQA can
be assumed as measuring the blur degradation of character/text. It is also observed that the gradient of the ideal character edge changes rapidly
while the gradient of
the degraded character edge has smooth change.
Inspired by the assumption and based on the observation, we use the gradient of character edge to measure the blur degradation of document images. To find meaningful patches containing character/text in document images, we choose maximally stable extremal regions (MSER) \cite{Matas2004Robust} as character patches, which is often used for character/text detection in OCR.

In this paper, 
to ensure that the legibility
of captured document images is sufficient for recognition,
we propose a no-reference DIQA framework, \textit{CG-DIQA}, based on
character gradient. Without need of prior training, the proposed method first employs the MSER based algorithm to detect significant character patches and then uses the gradient of character edge to describe the quality model of a document image.

\section{Approach}

In the proposed CG-DIQA method, we first convert an input document image 
to a grayscale image followed by downsampling to a specific size, then detect character candidates as selected patches, and finally compute the standard deviation of character gradients as the estimated quality scores for the document image. 
The flowchart of the
proposed method is demonstrated in Fig. \ref{fig:overview}.
Different steps of the CG-DIQA method are described in detail in the
consecutive subsections.

\subsection{Preprocessing}
To make quality assessment methods robust and efficient, preprocessing is generally required for DIQA.
In the preprocessing
step, a document image is initially converted into a grayscale image. 
Downsampling is also performed on the image in order to speed up the following processes if the resolution of document
images is greater than 1000$\times$1000.  
Smoothing is  unnecessary in the proposed method, since most of image noise will be avoided after extracting character patches and smoothing may deteriorate the blur degradation of document images.

\subsection{Character patch}
Before measuring document quality, it is necessary
to extract meaningful features for representing document images. 
Since the most significant feature of document images is character/text and DIQA is usually with respect to OCR performance,
we replace a document image with the patches that contain
characters during quality assessment. 
Using character patches can also make the proposed method more efficient, since it is easier for the method to handle patches rather than an
entire image.   

To extract character patches, the MSER based method \cite{Matas2004Robust} is first adpoted to detect character candidates, which perfroms well in scene text detection \cite{Alaei2015Document}.
The main advantage of the MSER based method is that such algorithm is able to find most legible characters even
when the document image is in low quality.
To remove repeating character candidates, the pruning process is incorporated
by minimizing regularized variations \cite{Alaei2015Document} in the MSER based method.

Since characters are often degraded to
smaller broken strokes which cause extremely lower/higher width-height ratio, the width-height ratio $\mathrm{r}_{c}$ of characters is also used to remove those broken strokes and meaningless non-character regions obtained with the MSER based method. $\mathrm{r}_{c}$ is set between  0.25 and 4 in this paper.   
The eventual bounding boxes represent typical character patches in document images.

\subsection{Character gradient}

It is observed that the gradient of degraded character edges is with smooth change.
Based on the observation, we use character gradient in character patches to measure the doument image degradation.

The image gradient can be calculated through
convolving an image with a linear filter. We have studied several often-used filters, such as the classic Sobel, Scharr and Prewitt filters, and find that the Sobel filter performs best in predicting quality scores of document images.
Thus, we choose the Sobel filter to calculate the character gradient in this paper. The Sobel filters on both directions are described as: 
\begin{equation}
\mathrm{f}_x=\left[
\begin{array}{ccc}
-1  & 0 & 1 \\
-2 & 0 & 2 \\
-1  & 0 & 1
\end{array}\right], \ \ 
\mathrm{f}_y=\left[
\begin{array}{ccc}
-1 & -2 & -1 \\
0 & 0 & 0 \\
+1& +2 & +1
\end{array}\right].
\end{equation}
Convolving $\mathrm{f}_x$ and $\mathrm{f}_y$ with a character patch (denoted by $\mathrm{c}$)
yields the horizontal and vertical gradient of the patch.
The gradient magnitude of patch $\mathrm{c}$ at position $(i, j)$, denoted by $\mathrm{m}_c(i, j)$, is computed as follows:
\begin{equation}
\mathrm{m}_c(i, j)=\sqrt{(\mathrm{c}\otimes \mathrm{f}_x)^{2}(i, j)+(\mathrm{c}\otimes \mathrm{f}_y)^{2}(i, j)}.
\end{equation}

In future, if better ways of calculating the character gradient emerge, it is easy to incorporate such ways into the proposed DIQA framework.

\subsection{Pooling strategy}
Borrowing the idea in the literatures \cite{Kumar2013Sharpness,Li2016No} where gradient features are used for DIQA, we compute the  overall quality score $\mathrm{s}$ of a document image as the standard deviation of character gradients via some pooling strategy. The widely used average pooling is adopted in this work to obtain the final
quality score for a document image. That is, first take the average $\mathrm{m}_a$
of character gradients, $$\mathrm{m}_a=\frac{1}{N}\sum_{c}\sum_{i,j}\mathrm{m}_c(i,j),$$ 
where $N$ is the total amount of pixels in all character patches,
and then compute the standard deviation of character gradients,
\begin{equation}
\mathrm{s}=\sqrt{\frac{1}{N}\sum_{c}\sum_{i,j}(\mathrm{m}_c(i,j)- \mathrm{m}_a)^{2}}.
\end{equation}
$\mathrm{s}$ is eventually used to describe the overall quality prediction score for a document image.

Since different character patches may contribute differently to the overall
quality score in a document image, the final quality score can be computed through weighting character gradients. Weighted pooling may have better DIQA accuracy than average pooling, but weighted pooling will result in more computing overhead and can make the pooling process more complicated. Furthermore, quality scores predicted with weighted pooling is more nonlinear, which is not beneficial to the following business process.

\begin{figure} 
	\centering 
	\subfigure[A low quality cropped document image ]{ 
		\label{fig:sample:a}
		\includegraphics[width=0.45\textwidth]{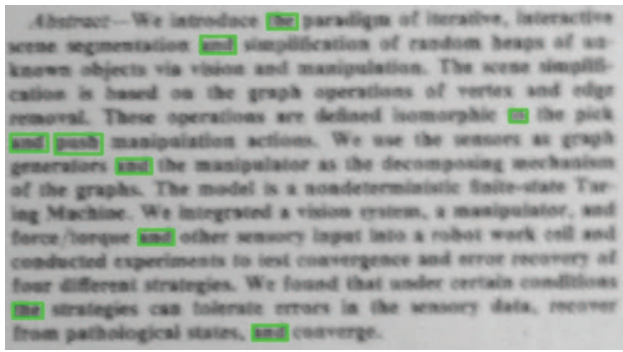}} 
	\subfigure[A degraded cropped document image ]{ 
		\label{fig:sample:b} 
		\includegraphics[width=0.45\textwidth]{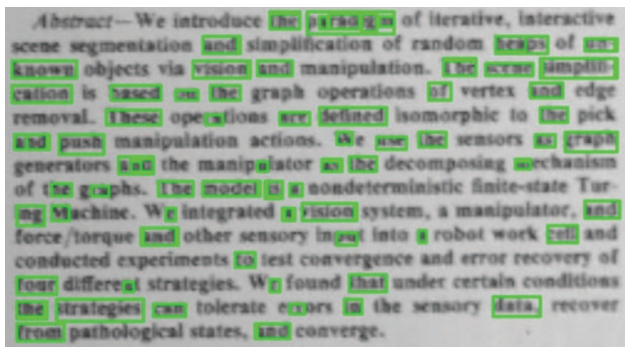}} 
	\subfigure[A high quality cropped document image]{ 
		\label{fig:sample:c} 
		\includegraphics[width=0.45\textwidth]{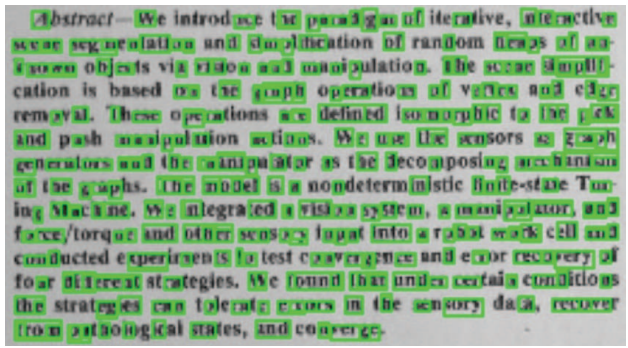}} 
	\caption{Samples of cropped document images in the DIQA dataset. The extracted character patches are surrounded with green bounding boxes. These three images are respectively with the average OCR accuracies of 33.62\%, 70.37\% and  93.01\%. Their quality prediction scores are 47.34, 61.45, and 72.21 respectively.} \label{fig:sample} 
\end{figure}

\section{Experiments}

\subsection{Dataset and evaluation protocol} 

We present evaluation results on a public DIQA dataset \cite{Kumar2013A}
containing a total
of 175 color images. These
images with resolution 1840$\times$3264 are captured from 25 documents containing machine-printed English characters using a smartphone. 6-8 photos were taken for each document to generate
different levels of blur degradations. In Fig. \ref{fig:sample}, we show three samples of cropped document images with different
degradation degrees from the DIQA dataset. Fig. \ref{fig:sample:a} is a low
quality image with severe blur that loses its readability. Fig. \ref{fig:sample:b} is a document image with the light degradation but it is still recognizable with human perceptual systems. And Fig. \ref{fig:sample:c} is a high quality image that can be easily read and recognized by OCR systems.

One traditional quality indicator for
document images is the OCR
accuracy \cite{Xu2016No}. Likewise, we define the OCR accuracy
as the ground truth for each document image in our experiments. Three OCR engines: ABBYY Fine Reader, Tesseract, and Omnipage, were run on each image in the dataset to obtain the OCR results. The OCR accuracy ranging from 0 to 1 was obtained through evaluating the OCR results with the ISRI-OCR evaluation tool.

To evaluate the performance of the proposed method, the predicted quality scores need to be correlated with the ground-truth OCR accuracies.
Thus, the Linear Correlation Coefficient
(LCC) and the Spearman Rank Order Correlation Coefficient
(SROCC) are used as performance indicators.

In our experiments, the LCC and SROCC are separately computed in a document-wise way. That is, for each document in the dataset, only its corresponding photos are taken into consideration while computing LCC or SROCC. Finally, we can get 25 LCCs and SROCCs involving the proposed method for this dataset. 
The medians of these 25 LCCs and SROCCs are used as the overall indicators for performance evaluation.
Since three OCR engines have huge difference in accuracy, to avoid that the evaluation results are overwhelmingly dependent on a certain OCR engine, we claim to use the average OCR accuracy of three engines for the evaluation purpose.

\subsection{Implementation and Results}
To get the optimal parameter setting for the proposed method, we tested different sets of the stable area size $s$ and the maximal variation $v_{max}$ in the MSER based method for extracting character patches. Parameter $s$ that is required to be neither too large nor small can play 
a role in eliminating the extracted false character regions. The maximal variation $v_{max}$ is used to prune the area that has similar size to its children. Experimental results show that the proposed method generally performs best when $v_{max}$ is set \textit{0.2} and $s$ is in a range \textit{13} pixels to \textit{0.001} of all pixels.

It is observed from our experiments that, once parameters are fixed in the MSER based method, the number of extracted character patches totally relies on the quality of document images. Although the extracted character patches are not too many for severely degraded document images, these patches, the most significant regions for character recognition, are enough to help estimate the quality score in a correct way.  

\begin{table}[t]
	\centering
	\caption{Results over the average OCR accuracy} \label{tab:comp:avg}	
	\begin{tabular}{l|c|c}
		\Xhline{1.2pt}		
		&Median LCC& Median SROCC\\
		\hline
		Sparse \cite{Peng2016Document}&0.935 &0.928 \\
		Proposed&\textbf{0.9841}&\textbf{0.9429}\\
		\hline
	\end{tabular}
\end{table} 

It is also worth noting that it is not the amount of character patches that plays important effects on the character gradient, but the quality of such patches on behalf of the entire document image. This strengthens that the character gradient can effectively reflect the document image quality. 

For example, in Fig.\ref{fig:sample}, there are only 48 character patches extracted in the severely degraded image with the average OCR accuray of 33.62\% (Fig.\ref{fig:sample:a}), 309 patches in the slightly degraded image with the average accuray of 70.37\% (Fig.\ref{fig:sample:b}), and 1615 patches in the high quality image with the average OCR accuray of 93.01\% (Fig.\ref{fig:sample:c}).
These three document images are respectively assessed to be with quality scores of
47.34, 61.45, and 72.21 using the proposed method.

We show our experimental results on the DIQA dataset in Table \ref{tab:comp:avg}. The median LCC and SROCC obtained with the proposed method are respectively 0.9841 and 0.9429. We compare the proposed method with the semi-supervised sparse representation based approach \cite{Peng2016Document} that computes the correlation coefficients as well in view of the average OCR accuracy.
As shown in Table \ref{tab:comp:avg}, the proposed method achieves the higher median
LCC and SROCC than the sparse approach \cite{Peng2016Document} that is based upon learning techniques.

\begin{table}[!ht]
	\centering
	\caption{Results on all document images} \label{tab:comp:all}	
	\begin{tabular}{l|c|c}
		\Xhline{1.2pt}		
		&LCC&SROCC\\
		\hline
		MetricNR\cite{Nayef2015Metric}&0.8867&0.8207 \\
		Focus\cite{Rusinol2014Combining} & 0.6467&N/A\\
		Proposed&\textbf{0.9063}&\textbf{0.8565}\\ 
		\hline
	\end{tabular}
\end{table}

To avoid the bias towards the good results in terms of the document-wise evaluation protocol, we also directly compute  one LCC (90.63\%) and one SROCC (85.65\%) over the average OCR accuracy
for all of the 175 document images in this dataset.
Table \ref{tab:comp:all} shows that our method performs much better than the other two metric-based methods: MetricNR \cite{Nayef2015Metric} and Focus \cite{Rusinol2014Combining}.

\subsection{Comparison}

To compare with other state-of-the-art quality assessment
approaches, we also compute the correlation values, LCC and SROCC, of the proposed method over three different OCR accuracies. 
In our experiments, seven general purpose DIQA approaches are selected for comparative anlaysis, including CORNIA \cite{Doermann2012Unsupervised}, CNN \cite{Kang2014A}, HOS \cite{Xu2016No}, Focus \cite{Rusinol2014Combining}, MetricNR \cite{Nayef2015Metric}, LocalBlur\cite{Chabard2015Local}, and Moments\cite{De2016Discrete}. 
Among them, the first three are based on learning techniques, while the others take advantage of
hand-crafted features. Since most of these methods either only focus one OCR accuracry or have no result on all document images, and it is hard to re-implement them to get optimal experimental results, we have to choose different methods with available accuracies for comparion. 

Table \ref{tab:comp:fr} illustrates the median LCCs and SROCCs of
six DIQA algorithms in terms of the FineReader OCR accuracy. From the results, we can see that, the coefficient LCC of the proposed method is slightly lower than three methods: MetricNR \cite{Nayef2015Metric}, CORNIA \cite{Doermann2012Unsupervised}, and HOS \cite{Xu2016No}. In addition, the proposed method is better than almost all other methods except the Focus method \cite{Rusinol2014Combining} in the SROCC coefficient. A nice, subtle highlight should be emphasized that the proposed method can perform well under both evaluation protocols, unlike other methods that can only work in a good way under a certain protocol.   
Since LCC can measure the degree of linear relationship between the predicted quality and OCR accuracy and SROCC can measure how well this relationship can be described using a monotonic function, experimental results demonstrate that the proposed CG-DIQA method is correlated with the FineReader OCR engine in a monotonic and linear way, which is more suitable for real applications.

In Table \ref{tab:comp:tsom}, we show the median LCCs and SROCCs of four approaches on the Tesseract and Omnipage accuracies. It can be observed that the proposed method provids a lot better OCR prediction scores than the
other three methods, even for different OCR engines.

\begin{table}[t]
	\centering
	\caption{Comparison over the FineReader OCR accuracy} \label{tab:comp:fr}	
	\begin{tabular}{l|c|c}
        \Xhline{1.2pt}			
		&Median LCC&Median SROCC\\
		\hline
		CORNIA\cite{Doermann2012Unsupervised}&0.9747&0.9286 \\
		CNN\cite{Kang2014A}&0.950&0.898\\
		HOS\cite{Xu2016No}&0.960 &0.909\\
		Focus\cite{Rusinol2014Combining}&0.9378&\textbf{0.96429}\\
		MetricNR\cite{Nayef2015Metric}&\textbf{0.9750}&0.9107\\
		Proposed&0.9523&0.9429\\
		\hline
	\end{tabular}
\end{table} 

	\begin{table}[!ht]
	\centering
	\caption{Comparison over the Tesseract and Omnipage OCR accuracies} \label{tab:comp:tsom}	
	\begin{tabular}{l|p{0.05\textwidth}|p{0.05\textwidth}|p{0.05\textwidth}|p{0.05\textwidth}}
		\Xhline{1.2pt}					
		\multirow{2}{*}{}&	\multicolumn{2}{c|}{Tesseract}	&\multicolumn{2}{c}{Omnipage}\\ 
		 \cline{2-5}
		&  Median LCC&   Median SROCC&   Median LCC&   Median SROCC\\	
		\hline		
		  LocalBlur\cite{Chabard2015Local}&N/A&0. 892&N/A&0. 725 \\
		 Moments\cite{De2016Discrete}&0.8197 &0.8207&N/A&0.6648\\
		 Focus\cite{Rusinol2014Combining}&0.9197&N/A&0.8794&N/A\\
		 Proposed&\textbf{0. 9591}&\textbf{0. 9429}&\textbf{0. 9247}&\textbf{0. 8295}\\
		\hline
	\end{tabular}
\end{table}

\section{Conclusion and Future Work}
In this paper, we propose a character gradient based method for document image quality
assessment. The method first extracts character patches via the MSER based method, then character gradients are computed for these patches, and finally the standard deviation of gradients are statistically obtained as the quality prediction score. In the method, it is assumed that character patches are more significant than widely-used square image patches in measuring the document image quality since they contain critical features for OCR. 
Our experimental results demonstrate that the proposed method can predict the quality score of document images very well in terms of both LCC and SROCC. The use of
some new techniques to extract features of
character patches
and a weighting strategy
based on patch size is our future research work.

\bibliographystyle{IEEEtran}

\bibliography{icpr}

\begin{thebibliography}{10}
\providecommand{\url}[1]{#1}
\csname url@samestyle\endcsname
\providecommand{\newblock}{\relax}
\providecommand{\bibinfo}[2]{#2}
\providecommand{\BIBentrySTDinterwordspacing}{\spaceskip=0pt\relax}
\providecommand{\BIBentryALTinterwordstretchfactor}{4}
\providecommand{\BIBentryALTinterwordspacing}{\spaceskip=\fontdimen2\font plus
\BIBentryALTinterwordstretchfactor\fontdimen3\font minus
  \fontdimen4\font\relax}
\providecommand{\BIBforeignlanguage}[2]{{%
\expandafter\ifx\csname l@#1\endcsname\relax
\typeout{** WARNING: IEEEtran.bst: No hyphenation pattern has been}%
\typeout{** loaded for the language `#1'. Using the pattern for}%
\typeout{** the default language instead.}%
\else
\language=\csname l@#1\endcsname
\fi
#2}}
\providecommand{\BIBdecl}{\relax}
\BIBdecl

\bibitem{Ye2013Document}
P.~Ye and D.~Doermann, ``Document image quality assessment: A brief survey,''
  in \emph{International Conference on Document Analysis and Recognition},
  2013, pp. 723--727.

\bibitem{Kang2014A}
L.~Kang, P.~Ye, Y.~Li, and D.~Doermann, ``A deep learning approach to document
  image quality assessment,'' in \emph{IEEE International Conference on Image
  Processing}, 2014, pp. 2570--2574.

\bibitem{Nayef2015Metric}
N.~Nayef, ``Metric-based no-reference quality assessment of heterogeneous
  document images,'' in \emph{SPIE Electronic Imaging}, 2015, pp.
  94\,020L--94\,020L--12.

\bibitem{Rusinol2014Combining}
M.~Rusinol, J.~Chazalon, and J.~M. Ogier, ``Combining focus measure operators
  to predict ocr accuracy in mobile-captured document images,'' in \emph{Iapr
  International Workshop on Document Analysis Systems}, 2014, pp. 181--185.

\bibitem{Chabard2015Local}
T.~Chabardès and B.~Marcotegui, ``Local blur estimation based on toggle
  mapping,'' in \emph{International Symposium on Mathematical Morphology and
  Its Applications to Signal and Image Processing}, 2015, pp. 146--156.

\bibitem{Matas2004Robust}
J.~Matas, O.~Chum, M.~Urban, and T.~Pajdla, ``Robust wide-baseline stereo from
  maximally stable extremal regions,'' \emph{Image \& Vision Computing},
  vol.~22, no.~10, pp. 761--767, 2004.

\bibitem{Alaei2015Document}
A.~Alaei, D.~Conte, and R.~Raveaux, ``Document image quality assessment based
  on improved gradient magnitude similarity deviation,'' in \emph{International
  Conference on Document Analysis and Recognition}, 2015, pp. 176--180.

\bibitem{Kumar2013Sharpness}
J.~Kumar, F.~Chen, and D.~Doermann, ``Sharpness estimation for document and
  scene images,'' in \emph{International Conference on Pattern Recognition},
  2013, pp. 3292--3295.

\bibitem{Li2016No}
L.~Li, W.~Lin, X.~Wang, G.~Yang, K.~Bahrami, and A.~C. Kot, ``No-reference
  image blur assessment based on discrete orthogonal moments,'' \emph{IEEE
  Transactions on Cybernetics}, vol.~46, no.~1, p.~39, 2016.

\bibitem{Kumar2013A}
J.~Kumar, P.~Ye, and D.~Doermann, \emph{A Dataset for Quality Assessment of
  Camera Captured Document Images}.\hskip 1em plus 0.5em minus 0.4em\relax
  Springer International Publishing, 2013.

\bibitem{Xu2016No}
J.~Xu, P.~Ye, Q.~Li, Y.~Liu, and D.~Doermann, ``No-reference document image
  quality assessment based on high order image statistics,'' in \emph{IEEE
  International Conference on Image Processing}, 2016, pp. 3289--3293.

\bibitem{Peng2016Document}
X.~Peng, H.~Cao, and P.~Natarajan, ``Document image quality assessment using
  discriminative sparse representation,'' in \emph{Document Analysis Systems},
  2016, pp. 227--232.

\bibitem{Doermann2012Unsupervised}
P.~Ye, J.~Kumar, L.~Kang, and D.~Doermann, ``Unsupervised feature learning
  framework for no-reference image quality assessment,'' in \emph{2012 IEEE
  Conference on Computer Vision and Pattern Recognition}, June 2012, pp.
  1098--1105.

\bibitem{De2016Discrete}
K.~De and V.~Masilamani, ``Discrete orthogonal moments based framework for
  assessing blurriness of camera captured document images,'' in
  \emph{Proceedings of the 3rd International Symposium on Big Data and Cloud
  Computing Challenges (ISBCC--16')}, V.~Vijayakumar and V.~Neelanarayanan,
  Eds., 2016, pp. 227--236.

\end{thebibliography}

\end{document}